\documentclass[lettersize,journal]{IEEEtran}
\usepackage{amsmath,amsfonts,amssymb}
\usepackage{algorithmic}
\usepackage{algorithm}
\usepackage{booktabs}
\usepackage{subcaption}
\usepackage{array}
\usepackage[caption=false,font=normalsize,labelfont=sf,textfont=sf]{subfig}
\usepackage{textcomp}
\usepackage{stfloats}
\usepackage{url}
\usepackage{verbatim}
\usepackage{graphicx}
\usepackage{cite}

\usepackage{multirow}
\usepackage{times}
\usepackage{epsfig}
\usepackage{cuted}
\usepackage{enumitem}

\usepackage{xcolor}

\begin{document}

\title{FlameFinder: Illuminating Obscured Fire through Smoke with  Attentive Deep Metric Learning
\thanks{H. Rajoli, S. Khoshdel, F. Afghah and X. Ma are with the  Holcombe Department of Electrical and Computer Engineering
        Clemson University, Clemson, SC, United States
        {\tt\small \{hrajolin, skhoshd, fafghah, xialom\}@clemson.edu.}}
\thanks{This material is based upon work supported by the Air Force Office of Scientific Research under award number FA9550-20-1-0090, the National Aeronautics and Space Administration (NASA) under award number 80NSSC23K1393, and the National Science Foundation under Grant Numbers CNS-2232048, and CNS-2204445.}
}

\author{Hossein Rajoli, Sahand Khoshdel, Fatemeh Afghah, Xiaolong Ma}





\maketitle

\begin{abstract}
FlameFinder, a novel deep metric learning (DML) framework, accurately detects RGB-obscured flames using thermal images from firefighter drones during wildfire monitoring. In contrast to RGB, thermal cameras can capture smoke-obscured flame features but they lack absolute thermal reference points, detecting many non-flame hot spots as false positives. This issue suggests that extracting features from both modalities in unobscured cases can reduce the model's bias to relative thermal gradients.
Following this idea, our proposed model utilizes paired thermal-RGB images captured onboard drones for training, learning latent flame features from smoke-free samples. In testing, it identifies flames in smoky patches based on their equivalent thermal-domain distribution, improving performance with supervised and distance-based clustering metrics. The approach includes a flame segmentation method and a DML-aided detection framework with center loss (CL), triplet center loss (TCL), and triplet cosine center loss (TCCL), to find the optimal cluster representatives for classification. Evaluation on FLAME2 and FLAME3 datasets shows the method's effectiveness in diverse fire and no-fire scenarios. However, the center loss dominates the two other losses, resulting in the model missing features that are sensitive to them. To overcome this issue, an attention mechanism is proposed making non-uniform feature contribution possible and amplifying the critical role of cosine and triplet loss in the DML framework. Plus, the attentive DML shows improved interpretability, class discrimination, and decreased intra-class variance exploiting several other flame-related features. The proposed model surpasses the baseline with a binary classifier by 4.4\% FLAME2 and 7\% in FLAME3 datasets for unobscured flame detection accuracy while showing enhanced class separation in obscured scenarios compared to fine-tuned VGG19, ResNet18, and three other backbone models tailored for flame detection.

\end{abstract}

\vspace{-0.3cm}
\section{Introduction}

Forest fires cause property damage, injuries, and deaths, harming people and the economy. Extreme wildfire events pose greater risks and challenges, therefore efficient wildfire detection and management are essential to mitigate these impacts. \cite{tedim2018defining}. Existing technologies are not yet capable of accurate and timely detection of wildfire noting the limited lifetime and sensing range of thermal/smoke sensors, delays and low spatial resolution associated with satellites' observations \cite{afghah2019wildfire}. In the context of wildfire detection, UAVs have emerged as a promising technology for wildfire detection and management. UAVs offer high mobility, flexibility, low deployment cost, and real-time data collection capabilities, making them well-suited for monitoring wildfires in remote and challenging terrains \cite{SHAMSOSHOARA2021108001}. Deploying UAV systems for  wildfire management has not been limited to wildfire detection \cite{BOROUJENI2024102369}.  Although the performance of UAVs in other tasks such as active wildfire monitoring with decision-making systems \cite{khoshdel2024pyrotrack} and spread modeling of fire frontiers, all rely on how accurate their input is provided by a flame detection system. While many improvements have been done in terms of genreal flame detection, commercial UAVs with visual cameras cannot yet accurately identify the flame locations when smoke and water vapor block visible-light spot fires, resulting in false negatives. It should be noted that while smoke detected in RGB images can be an indicator of fire, it cannot accurately pinpoint the fire location and intensity due to the complex dynamics of smoke, wind and fire and several other effective variables. As a result, utilizing thermal images were hypothesized as a solution for drone-based flame mapping \cite{gaur2020video}.\vspace{0.1cm}

While some studies focus on thermal-domain fire detection images with exploring sensor response and image processing techniques, Flame detection for smoke-obscured samples is understudied and remains a major gap this area \cite{sousa2020thermal}. On the other hand the lack of absolute thermal reference points and the existence of other thermal sources make models trained solely in thermal domains inaccurate. Shortcomings of pure RGB and pure thermal domains methods pose the fact that the problem of obscured flame detection naturally needs complementary modalities. Following this idea, to extract features associated with flames and not just hotspots occurring in the wildland (e.g. carried smoke), an idea would be to to transfer knowledge (by means of labels), from unobscured samples in the RGB domain (as an indirect supervision) to the unannotated target thermal domain. Deep neural networks (DNN) have shown potential for UAV-based flame detection by integrating data from RGB and thermal cameras \cite{Chen}. Such integration provides complementary information, improving the discrimination of flames from other heat sources. Transfer learning and fine-tuning on DNN architectures have demonstrated enhanced flame detection accuracy in UAV-collected images \cite{treneska2021wildfire}. This process can be initiated by providing annotations with a segmentation algorithm on the source domain. Next, to aggregate the embedded information of the thermal domain, a model is trained to shape a compressed latent space with objectives that capture various flames, rather than simply getting biased towards the thermal domain intensity. Solving this challenge is core to overcome the problem of relative thermal reference points and non-flame hotspots appearing as false positives.

Among representation learning approaches proposed for unsupervised clustering of categories, deep metric learning (DML) has shown major success for learning similarity and distance metrics by combining deep learning with distance metric learning \cite{kaya2019deep}. Regarding this idea, given the rich latent space constructed with DML captures enough features from both classes on the thermal domain, flames that are obscured by smoke in the RGB domain will have corresponding thermal representations that fall in the positive (flame) class of the discriminator. Moreover, samples with no flames that have similar smoke patterns (local RGB patterns) to occluded flames, won't appear similar to flames in the thermal domain. \vspace{0.1cm}

Our proposed work uses bi-modal data in RGB and thermal domains such that one modality (RGB) is used for sample annotation, while the other modality (Thermal) offers latent features associated with the sample's classes, making inference possible with merely the second modality (Thermal), and providing flame predictions for obscured areas. Such tasks fall under the umbrella of domain adaptation, where knowledge is transferred from one domain to the other. \cite{farahani2021brief} Here, it should be mentioned that discrepancies between the source and target domain distributions can affect the performance of DML models on multi-modal data \cite{7478033, 7529190}. \vspace{0.1cm}
This study employs Distance Metric Learning (DML) to enhance the model's performance in separating class clusters within the embedding space, specifically for detecting smoke-covered flames using paired RGB-thermal images. To facilitate the generation of data patches for fire spread modeling and monitoring, we introduce a segmentation and annotation framework (\ref{unobscured flame segmentation}). This framework labels a patch as "flame" if it contains at least one flame-annotated pixel and as "no-flame" otherwise. While many advanced domain adaptation methods offer innovative solutions to align features of the complementary domains, our simple indirect supervision approach benefits from the pairwise nature of patches in source and target domains to make the convergence path smoother than unsupervised cases. After annotation is provided, a DML method is utilized for the annotated data, training a feature extractor to distinguish between flame and non-flame patches. A classifier is then employed to predict flames in smoke-obscured areas where no labels are available. The framework incorporates three loss functions in the DML framework to learn an optimal embedding function in the latent space. Recognizing the dominance of center loss, particularly for magnitude-sensitive features, we propose an attention mechanism to balance feature contributions across the three DML loss gradients, harnessing the full potential of the constructed embedding space. This feature-selective loss smoothing enhances class discrimination in the latent feature space, addressing the issue of loss domination in multi-objective optimization. The main contributions of this work include:
\vspace{0.1cm}
\begin{itemize}
\setlength\itemsep{0.2cm}

    \item An improved thermal-based flame detection method onboard aerial vehicles using a clustering-guided feature representation learning over RGB and thermal images and domain adaptation in training to improve performance and reliability,

    \item Proposing learnable class representatives optimized with SGD, resulting in enhanced discrimination of class-related latent features

    \item Proposing a novel attention mechanism to balance latent feature contribution in DML loss functions, resulting in diverse feature discovery.


\end{itemize}

\section{Related Works}
\label{sec:rel_work}

Supervised machine-learning techniques are increasingly applied in wildfire detection. In \cite{chen2019uav}, an approach combines local binary pattern (LBP) and SVM for smoke detection from UAV-captured RGB images, and a CNN for flame detection with preprocessing steps such as histogram equalization and low-pass filtering. XtinguishNet \cite{sethuraman2022idrone} trains a CNN on IoT data for real-time wildfire detection, offering weather insights and fire intensity estimates. \cite{nguyen2021visual} presents a MobileNet-based single-shot detection (SSD) model for real-time UAV-based wildfire monitoring. Some other works employ heuristic masks such as using HSV domain filtering for flame-sensitive features \cite{ryu2021flame}. Some other works focus on Object-detection-based approaches with a simplified YOLO model \cite{Shen} or MobileNet combined with YOLO-v4 \cite{li2021lightweight}. Notably, none of these works explore multi-modal input for obscured flame detection.

\subsection{DML-based Object Detection/Classification}
DML-based object detection has become popular in computer vision for learning distance metrics as a path to learn rich feature representations \cite{kaya2019deep}. In \cite{karlinsky2019repmet}, the proposed method tunes the backbone network parameters to find an embedding function for multi-modal data in a single training process. They use a subnet architecture for embedding function training and a multi-modal mixture distribution for class posterior computation. In facial expression recognition, \cite{rajoli2023triplet} improves classification accuracy using DML and triplet loss, with stochastic gradient descent (SGD) optimizing class prototypes, backbone, and classifier parameters together. DML's popularity in few-shot learning is evident in \cite{hong2022few}, employing representation generation, and distance estimation modules beside DML for object representation, classification, and location estimation. \cite{lu2022decoupled} discusses Decoupled Metric Network (DMNet) for single-object detection, introducing Decoupled Representation Transformation (DRT) and Image-level Distance Metric Learning (IDML) to address representation disagreement and enhance generalization. DML-based methods excel in multi-modal detection for capturing correlations and complex features across modalities \cite{7244184}. However, challenges arise in handling high-dimensional feature spaces and maintaining a balance between dimensionality size, and the model performance. The density of the embedding space influences generalization in DML-based models \cite{pmlr-v119-roth20a}, while significant dimension reduction may impact performance. Although various works contribute to enhancing DML-based approaches, this work pioneers the use of DML in flame detection, introducing an aggregated loss including triplet and cosine losses with center loss to emphasize class discrimination aided by attention.

\subsection{Attention in Deep Metric Learning}

Traditional DML models effectively learn embedding functions but struggle with dynamic focus on different aspects of the data and adapting to variable-length inputs \cite{bahdanau2014neural}. Attention mechanisms in DML, as highlighted by \cite{kim2018attention}, enhance feature embeddings and improve performance on image retrieval tasks by allowing each learner to attend to different parts of the input. \cite{wang2019deep} employs attention for feature balancing and class representation weighting, introducing Class-Aware Attention (CAA) to identify noisy images and improve convergence in DML. Addressing DML challenges, \cite{kotovenko2023cross} introduces a cross-attention mechanism between image embeddings, establishing a hierarchy of conditional embeddings to enhance tupled representations. \cite{seidenschwarz2021learning} goes beyond pairs and triplets in DML, proposing message-passing networks with an attention mechanism to weigh the importance of each neighbor in a mini-batch. \cite{li2020unsupervised} introduces transformed attention consistency and contrastive clustering loss for robust similarity measures in DML. In several application domains, attention shows significant improvement for compressed feature embedding. In \cite{yang2020deep}, mixed attention mechanisms aid histopathological image retrieval. \cite{dong2021deep} addresses the challenge of varying input sizes in DML with a hybrid channel and spatial attention approach for pavement distress classification. \cite{coskun2018human} uses an attentive RNN in human motion analysis to measure the similarity between motion patterns. Joint embedding of visual modalities, such as optical flows and RGB, is explored in \cite{wang2019deep} for near-duplicate video retrieval, employing attention modules to capture different levels of granularity in video-level representation.

\begin{figure*}[htb]
\centerline{\includegraphics[width=2\columnwidth]{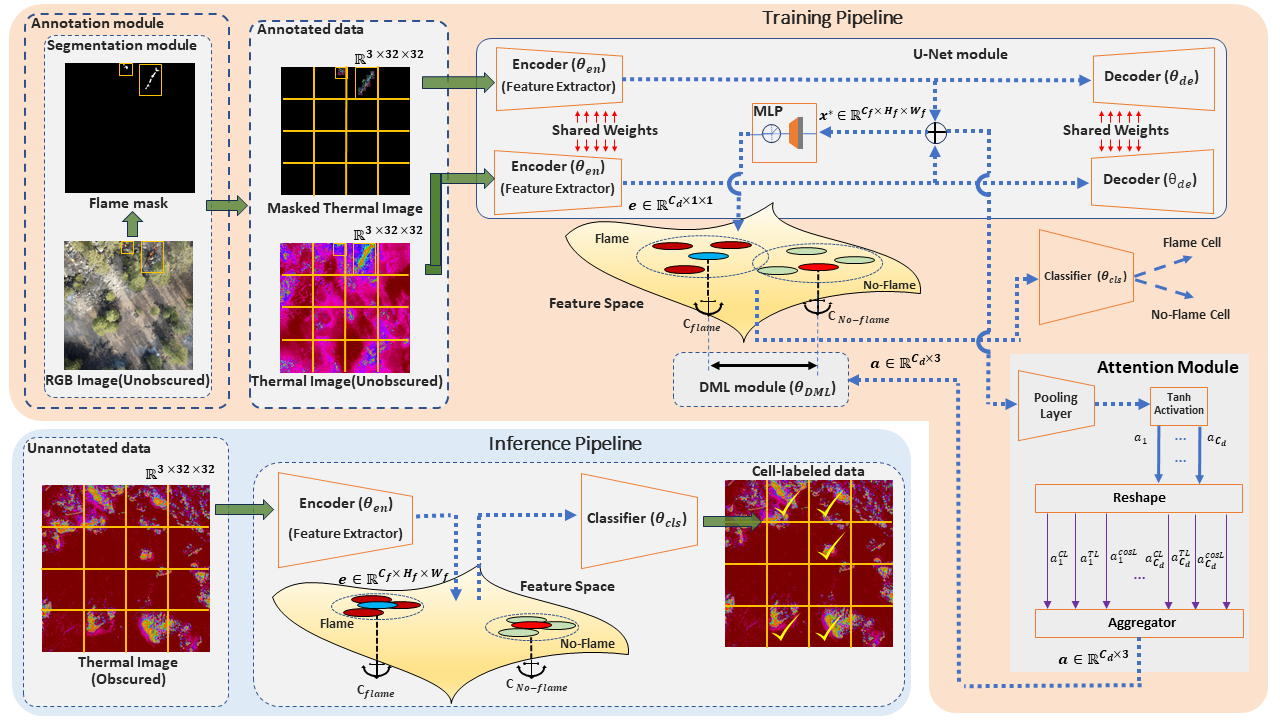}}
\captionsetup{justification=justified}
\caption{ (a) \textbf{Training}: simultaneously learning the encoder, decoder, embedding space, and classifier. The class's prototype and the embedding function are jointly trained, while each feature's loss contribution is balanced with the attention module. Finally, the thermal patches are predicted as flame/no-flame; (b) \textbf{Inference}: Obscured thermal images are introduced to the detector, and based on the classifier's output the corresponding cluster (patch label) is predicted}.
\label{main idea}

\end{figure*}

\section{Proposed Method}\label{sysmodl}
\label{sec:sys_modl}

This section describes our approach for flame detection and classification (Fig. \ref{main idea}), aiming to identify flames obscured by smoke layers, especially in thermal images. The objective is to learn the relationship between flame patterns in RGB and their corresponding thermal representation. The process involves three main steps: \vspace{0.3cm}\\
\textbf{\textit{1. Unobscured Flame Annotation:}} Annotating visible flames in RGB images using Alg. \ref{unobscured flame segmentation}. \vspace{0.3cm}\\
\textbf{\textit{2. DML-guided Learning:}} Training an Auto-Encoder and Classifier on annotated thermal images with DML for flame detection (Alg.\ref{proposed instruction}). \vspace{0.3cm}\\
\textbf{\textit{3. DML-guided Inference:}} Using the trained DML-based model to detect flames in obscured thermal patches. An attention mechanism balances the contribution of utilized loss functions in all features of the latent space.

\vspace{-0.2cm}
\subsection{Unobscured Flame Annotation}
\label{unobscured flame segmentation}

In response to the scarcity of annotated RGB-thermal image pairs for wildfire segmentation, we propose a tailored filtering approach for unobscured flame detection in top-down UAV-captured RGB images. The RGB-unobscured flame segmentation algorithm (Alg. \ref{unobscured flame segmentation}) creates a segmentation mask based on heuristic channel value inequalities, which can be later value. Next, smoothing and pixel-removal operations are applied, and resulting patches are labeled as 'flame' or 'no-flame,' providing indirect supervision for learning to detect flames in unannotated patches during the inference phase. This strategy aims to overcome challenges associated with the lack of absolute thermal reference points, mitigating misclassification in thermal images due to relative temperature gradients. It is worth highlighting the fact that this simple, yet automated segmentation is prone to error, and as a result, label noise may be limiting the performance of the model in terms of the classification accuracy. A comparative analysis on the effect of segmentation noise is done in \ref{annotation analysis} to further justify the automated segmentation.  \\



\subsubsection{\textbf{RGB Mask Generation}} 
\begin{figure}[htb]
\centerline{\includegraphics[width=0.9\columnwidth]{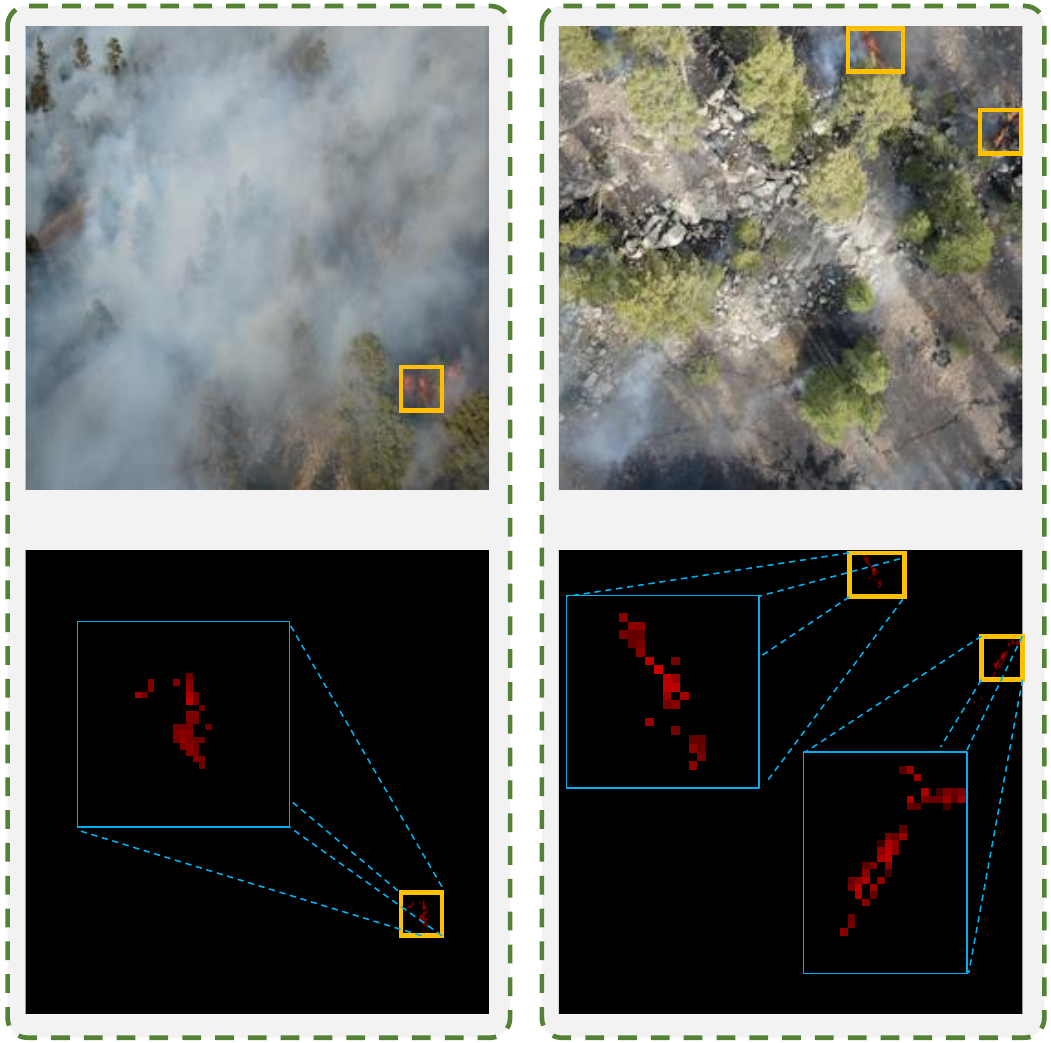}}
\captionsetup{justification=justified}
\caption{\textbf{Unobscured flame segmentation.} sample images demonstrating the performance of the proposed unobscured flame segmentation method in identifying flame instances covered by smoke. }
\label{RGB_mask}
\end{figure}

In light of the approaches proposed by \cite{buza2022unsupervised, dzigal2019forest} for forest fire detection, we have developed a well-fitted masking method to cater to the specific type of fire encountered in real-world situations, namely, spot flames captured with a top-down perspective by UAVs. Alg.~\ref{unobscured flame segmentation} explains our approach for detecting unobscured flames in RGB images. This algorithm sets a criterion based on channel means, values, and inter-channel distances to generate a flame segmentation mask. The combination of three masks representing RGB criteria, reflecting scene-specific features such as the dominant green background in forest fire images, facilitates the identification of flames amidst the smoke. We calculate a flame segmentation mask by combining three masks, each representing an RGB criterion for candidate flames. The reason for choosing such criteria lies in some scene-specific features, such as the dominant green texture of the background in forest fire images. In addition, the combination of green and red channels creates orange and yellow tones that are the dominant colors in the observed flames. Fig \ref{RGB_mask}, demonstrates the segmentation results followed by smoothing and pixel-removal operations. \\

\subsubsection{\textbf{Thermal Patch Labeling}}  Eventually, any patch containing part of a detected flame in the output image will be labeled positive as 'flame', and all other patches will be labeled negative as 'no-flame'. All equivalent thermal patches of the pair thermal image will be annotated likewise. To demonstrate the effectiveness of the algorithm, a set of such patches are pixel-annotated by a human expert, and the corresponding results are presented in Section 4.2. It should be noted that the mask is not used as an annotation module in the inference phase. We intend to classify smoky patches as flame/no-flame where no reliable class is provided. Thus, no labels are given or produced for the inference phase. In other words, the mask provides indirect supervision for a dataset of unannotated patches, in which the associated side-by-side thermal features of flames are learned to detect flames for classification on smoky patches in the inference phase. This indirect thermal patch labeling aims to avoid using the relative thermal spectrum captured in thermal images. This issue in thermal imagery is rooted in the lack of absolute thermal reference points. As a result, thermal values indicate the relative temperature of an area compared to its surroundings. It should be noted that in the case of using thermal images directly, any other local temperature gradient can lead the patch to be misclassified as a flame. In conclusion, the joint usage of RGB and thermal modalities in segmentation and classification respectively surpasses the discussed issues raised in uni-modal RGB/thermal detection It is noteworthy that for optimal segmentation performance, the threshold values ($\tau_g$, $\tau_b$) and parameters ($\alpha$, $\beta$) need to be fine-tuned to accommodate for misty flames arising due to smoke.

\subsubsection{\textit{\textbf{Comparative Annotation Analysis}}}
\label{annotation analysis}
For a set of 8640 patches, originating from 135 images, the proposed annotation mechanism \ref{unobscured flame segmentation} was able to identify 81\% of the flames detected by an expert. One main concern would be the effect of mislabeled patches on the final performance of the model. To be more specific, when a model is trained with noisy annotated data because the inference phase performance will also be measured with noisy labels, the annotation noise will be hindered by the good results, while the model may not perform well in reality. To address this concern and measure the annotation noise effect, a comparison between the inference phase performance should be done between the two cases of expert-annotated and automated annotated data. As training is impractical with expert-annotated data due to the large volume of required data, we have conducted a simple experiment, where we tested the fine-tuned model (on automated annotation) on the expert-labeled data. Details of the performance are shown in the confusion matrix presented in Table~\ref{tab:confusionmatrix}. Class imbalance is due to the very few number of flame patches extracted from no-flame images, whereas many no-flame patches are present in flame-labeled images. As expected, the accuracy decreases on expert-annotated data as the model predicts several instances of flames (mis-annotated due to annotation noise) as 'no-flame' and vice versa. \vspace{0.2cm}

 Following the performance degradation on expert-annotated data in terms of the classification accuracy (82.9\% in Table \ref{tab:confusionmatrix} , compared to 97.8\% in Table \ref{tab:AchievementComparison}), the following facts should be kept in mind when utilizing an the automated annotation algorithm proposed in Alg. \ref{unobscured flame segmentation}: 
 \vspace{0.2cm}
\begin{enumerate}
    \item 

    Some flames may be hindered by an opaque object such as a tree branch. Despite here we were limited to the currently available datasets, future works may use consecutive frames of a video, multi-view imaging or other techniques to reduce the chances of object obfuscation. 

    \item  Although the expert annotation is more accurate, the fact that manual annotation of the whole dataset is time-consuming due to large number of final patches, prevents us from a more comprehensive comparative study between manual and automated annotation. 

    \item In some cases, the magnitudal spatial correlation between thermal and RGB patterns, and also the correlation between thermal patterns in obscured vs. unobscured patches may be reduced due to various reasons such as: smoke mobility, proximity of smoke to the camera, etc. This label noise may increase the misclassified instances, as seen in Fig \ref{mislabled_samples}. However most of such cases are unavoidable while training and should be detected and controlled with equipment in the data collection phase (e.g. utilizing more modalities in field can prevent the model learn noisy associations and boost the inference accuracy.) 

    \item There may be algorithms with much higher accuracy than the proposed heuristic approach, but this approach suits the real-time implementation limitations for computational efficiency as it only serves as the pre-processing (annotating) part of the design.

\end{enumerate}

\noindent
\begin{minipage}{\linewidth}
    \noindent\rule{\linewidth}{0.7pt} 
    \noindent\textbf{Algorithm FlameSeg (Alg.~\ref{unobscured flame segmentation}) :} RGB-unobscured flame segmentation
using image processing \\
    \vspace{-0.2ex}
    \noindent\rule{\linewidth}{0.7pt} 
    
    \vspace{0ex} 
    
    \textbf{Inputs:}
    \begin{itemize}[itemsep=1pt, topsep=0pt, label=\textbullet, font=\small, leftmargin=4ex]
        \item $Img_{[R,G,B]}$ (input images),
        \item $R_m$, $G_m$, $B_m$ (red, green, blue channel means),
        \item $\tau_g$, $\tau_b$ (green and blue channel value threshold),
        \item $\alpha$, $\beta$ (red-to-green low and high margin coefficients)
    \end{itemize}
    
    \vspace{0.2cm} 
    
    \textbf{Output:} \,\, Flame segmentation mask $m$.
    
    \vspace{0.2cm} 
    
    \begin{enumerate}[label=\arabic*., itemsep=1pt, topsep=1pt, font=\small, leftmargin=*]
        \item filtering based on the distance to the mean:
            \vspace{0.1cm}

            \begin{flushleft}
            $
            {m_1(x,y)} \,\,
            {\longleftarrow} \,\,
            \left\{
            \begin{aligned}
            &0 \, , \,\,\,\text{$Img_{[R]} - R_m < Img_{[B]} - B_m$}\\
            &0 \, , \,\,\,\text{$Img_{[R]} - R_m < Img_{[G]} - G_m$}\\
            &1 \, , \,\,\,\text{Otherwise}
            \end{aligned}
            \right.
            $
            \end{flushleft}

        \vspace{0.3cm}
        \item filtering based on the value:
            \vspace{0.1cm}
            \begin{flushleft}
            $
            {m_2(x,y)} \,\,
            {\longleftarrow} \,\,
            \left\{
            \begin{aligned}
            &0 \, , \,\,\,\text{$Img_{[G]} > \tau_g$}\\
            &0 \, , \,\,\,\text{$Img_{[B]} > \tau_b$}\\
            &1 \, , \,\,\,\text{Otherwise}
            \end{aligned}
            \right.
            $
            \end{flushleft} 
        \item filtering based on the inter-channel distance:
            \vspace{0.1cm}
            \begin{flushleft}
            ${m_3(x,y)} \,\,
            {\longleftarrow} \,\,
            \left\{
            \begin{aligned}
            &0 \, , \,\,\,\text{$|Img_{[R]} - Img_{[G]} | \leq \alpha Img_{[G]}$}\\
            &0 \, , \,\,\,\text{$|Img_{[R]} - 2*Img_{[G]} | \geq \beta Img_{[R]}$}\\
            &0 \, , \,\,\,\text{$|Img_{[G]} - 2*Img_{[B]} | \leq \delta Img_{[G]}$}\\
            &1 \, , \,\,\,\text{Otherwise}
            \end{aligned}
            \right.
            $
            \end{flushleft}
            
        \vspace{0.1cm}
        \item applying the masks, removing single pixels, and
smoothing
        
    \end{enumerate}
    
    \vspace{-1ex} 
    
    \noindent\rule{\linewidth}{0.7pt} 
\end{minipage}
\vspace{1ex} 

\begin{table}[h]
\centering
\caption{Confusion Matrix}
\label{tab:confusionmatrix}
\begin{tabular}{cccccc}
\toprule
& & \multicolumn{3}{c}{\textbf{Predicted}} & \\
\cmidrule{3-5}
& & \textbf{Flame} & \textbf{No-Flame} & \textbf{Total} \\
\midrule
\multirow{2}{*}{\textbf{Actual}} & \textbf{Flame} & 2093 & 739 & 2832 \\
& \textbf{No-Flame} & 643 & 4625 & 5268 \\
\midrule
& \textbf{Total} & 2736	& 5364 & 8100 \\
\midrule
\multicolumn{2}{c}{\textbf{Precision (Flame)}} & \multicolumn{3}{c}{76.5 \%} \\
\multicolumn{2}{c}{\textbf{Recall (Flame)}} & \multicolumn{3}{c}{73.9 \%} \\
\multicolumn{2}{c}{\textbf{Accuracy (Flame)}} & \multicolumn{3}{c}{\textbf{82.9 \%}} \\
\bottomrule
\end{tabular}
\end{table}

\color{black}
\vspace{-0.4cm}
\subsection{DML Loss Derivatives}
\label{derivatives}
Here the derivatives of the loss functions used in DML losses are shown. (Equations \ref{der_to_e_TL}-\ref{der_to_a_CL})
\begin{flushleft}
\begin{align}
\label{der_to_e_TL}
\frac{\partial \mathcal{L}_{TL}}{\partial e} = \frac{1}{m}
\sum_{x^{(i)} \in \mathcal{K}}^{} \,\sum_{j=1}^{c_d} {a^{TL}_{ij} \big( (c^{n}_{y^{(i)}{j}})+ (c^{p}_{y^{(i)}{j}}) } \big) H(\mathcal{L}_{TL})
\end{align}
\end{flushleft}

\begin{flushleft}
\begin{align}
\label{der_to_a_TL}
\frac{\partial \mathcal{L}_{TL}}{\partial a_{TL}} = \,\Big(& \, \frac{1}{2m} \sum_{i=0}^{m} \sum_{j=1}^{c_d} {\,(\,\| {(e^{(i)}_{j})} - (c^{p}_{y^{(i)}{j}}) \| _{2}^{2}} \nonumber\\ 
& -{\| ({e^{(i)}_{j}}) - (c^{n}_{y^{(i)}{j}}) \| _{2}^{2}} + \alpha)\,\Big) H(\mathcal{L}_{TL})
\end{align}
\end{flushleft}

where $H(\mathcal{L}_{TL})$ represents the Heaviside step function. It should be noted that in Eq.~\ref{der_to_e_TL} to Eq.~\ref{der_to_a_CL},  $a^{TL}_{ij}$, $a^{CL}_{ij}, e^{(i)}_{j}, c^{p}_{y^{(i)}{j}}, c^{n}_{y^{(i)}{j}}$, are scalars, whereas $a^{cosL}_{i}$, $e^{(i)}$, $c^{n}_{y^{(i)}}$, and $c^{p}_{y^{(i)}}$ are vectors with length $c_{d}$. Plus, the $"\odot"$ and  $"\cdot"$ symbols used in Eq.~\ref{der_to_e_TL}, Eq.~\ref{der_to_a_TL}, and Eq.~\ref{der_to_e_cos} denote element-wise multiplication and inner product operations respectively.
 \vspace{-0.1cm}

 \vspace{-0.4cm}
\begin{align}
\label{der_to_e_cos}
&\frac{\partial \mathcal{L}_{cosL}}{\partial e} = \nonumber\\
&\frac{-1}{2m}\sum_{i=1}^{m} a^{cosL}_{i} \odot \Bigg( \Big(\frac{ c^{p}_{y^{(i)}}}{\big\lVert e^{(i)}\big\rVert \big\lVert c^{p}_{y^{(i)}}\big\rVert} - \frac{(e^{(i)}\cdot c^{p}_{y^{(i)}})e^{(i)}}{\big\lVert e^{(i)}\big\rVert^3 \big\lVert c^{p}_{y^{(i)}}\big\rVert}\Big) \nonumber\\
&- \Big(\frac{c^{n}_{y^{(i)}}}{\big\lVert e^{(i)}\big\rVert \big\lVert c^{n}_{y^{(i)}}\big\rVert} - \frac{(e^{(i)}\cdot c^{n}_{y^{(i)}})e^{(i)}}{\big\lVert e^{(i)}\big\rVert^3 \big\rVert c^{n}_{y^{(i)}}\big\rVert}\Big) \Bigg)
\end{align}
 \vspace{-0.4cm}

\begin{flushleft}
\begin{align}
\label{der_to_a_CosL}
\frac{\partial \mathcal{L}_{CosL}}{\partial a_{cosL}} = \, \, & - \, \frac{1}{2m} \sum_{i=0}^{m} \sum_{j=1}^{c_d} \,\Big( \frac{(e^{(i)}_{j} c^{p}_{y^{(i)}{j}})}{\left\lVert (e^{(i)}) \right\rVert \big\lVert (c^{p}_{y^{(i)}}) \big\rVert} \nonumber\\
&- \frac{(e^{(i)}_{j}) (c^{n}_{y^{(i)}{j}})}{\left\lVert (e^{(i)}) \right\rVert \big\lVert (c^{n}_{y^{(i)}}) \big\rVert} \Big)
\end{align}
\end{flushleft}

\begin{flushleft}
\begin{align}
\label{der_to_e_CL}
\frac{\partial \mathcal{L}_{CL}}{\partial e} = \frac{1}{m}
\sum_{i=1}^{m} \,\, \sum_{j=1}^{c_d} a^{CL}_{ij} {\big( ({e^{(i)}_{j}})- (c_{y^{(i)}{j}}) } \big)
\end{align}
\end{flushleft}
 \vspace{-0.4cm}

\begin{flushleft}
\begin{align}
\label{der_to_a_CL}
\frac{\partial \mathcal{L}_{CL}}{\partial a_{CL}} = \frac{1}{2m}
\sum_{i=1}^{m} \,\, \sum_{j=1}^{c_d} \| {(e^{(i)}_{j})} - (c_{y^{(i)}{j}}) \| _{2}^{2}
\end{align}
\end{flushleft}

\subsection{DML-guided Learning} 
During the training phase (Alg.~\ref{proposed instruction}), it is crucial to sample inputs from smoke-free images to ensure a comprehensive exploration of flame representations in the thermal domain.
We oversample the flame patches to address the class imbalance in the dataset and weigh the loss accordingly. Our final dataset comprises cropped patches from thermal images labeled as either "flame" or "no-flame." \vspace{0.1cm} 

The proposed framework comprises an encoder and decoder (based on the U-net autoencoder architecture \cite{ronneberger2015u}), a classifier, and a DML module that work together to learn features of the input data through a multi-task loss function. The loss function is a weighted sum of three terms: cross-entropy classification loss, autoencoder reconstruction loss, and the DML losses described afterwards. Let $D_m = \{ (x^{(i)}, y^{(i)}) \mid i = 1, 2, \ldots, m \}$ be a mini-batch of $m$ samples, where $x^{(i)} \in \mathbb{X}$ represents pairs of thermal patches and their corresponding flame masks from the training set, and $y^{(i)} \in \{1, 2\}$ is the corresponding categorical label (flame or no-flame). At the output of the encoder, the sample representation is shown by $x^{\ast} \in \mathbb{R}^{C_{f} \times H_{f} \times W_{f}}$, where $C_{f}$, $H_{f}$, and $W_{f}$ are the number of channels, rows, and columns of the encoded samples, respectively. A convolutional layer and a global average pooling layer follow the encoder. The encoder maps the encoded samples of the mini-batch to their feature space representation, denoted by $e \in \mathbb{R}^{C_{d} \times 1 \times 1}$, where $C_{d}$ is the number of channels (the length of the feature embedding vector).\vspace{0.1cm} \\

A fully connected (FC) layer in the classifier network maps the embedding to logits of $x^z \in \mathbb{R}^{2}$. The Softmax function is applied to obtain a probability distribution $Pr(y = j \mid x^{(i)})$. The discrepancy between the predicted label $\hat{y}$ and the true label $y^{(i)}$ is calculated using the cross-entropy (CE) loss function, denoted by $\mathcal{L}_{BCE}$, as follows:\vspace{-0.2cm}
\begin{equation}
\label{CE_loss}
\mathcal{L}_{BCE} = -\frac{1}{m} \sum_{i = 1}^{m} \sum_{j = 1}^{2} y^{(i)} \log Pr(y^{(i)} = j \mid x^{(i)}).
\vspace{-5pt}
\end{equation} 
By implying the reconstruction loss, we also ensure the latent representation learned by the encoder, which is core to classifying and clustering, contains the essential features of the input domain, preserving the input information as much as possible. \vspace{-0.2cm}
\begin{equation}
\label{rec_loss}
\mathcal{L}_{rec} = \frac{1}{m}\sum_{i=1}^{m} \left\lVert x^{(i)} - f(g(x^{(i)}))\right\rVert^2
\vspace{-5pt}
\end{equation}
where $m$ is the number of examples in the dataset, $x^{(i)}$ is the $i$-th input example, $g$ and $f$ are the encoder and decoder.
\vspace{0.1cm}

\subsubsection{\textit{\textbf{Multi-task Loss Function}}}
 The multi-task loss function aims to optimize multiple objectives simultaneously. This loss function includes the direct cross-entropy ($\mathcal{L}_{BCE}$) loss for classification, the reconstruction loss for an autoencoder  ($\mathcal{L}_{rec}$), and the distance loss for DML  ($\mathcal{L}_{DML}$), which considers the distance of each sample's latent representation to the cluster representatives.
\begin{equation}
\label{multi_task_loss}
\mathcal{L} = \mathcal{L}_{BCE} + \lambda \, \mathcal{L}_{rec} + \, \mathcal{L}_{DML},
\end{equation}


\subsubsection{\textit{\textbf{DML Metrics}}}
\label{DML Metrics}
Triplet loss optimizes anchor, positive, and negative sample distances, along with emphasizing discriminative features, while cosine similarity is based on feature vector angle suiting high-dimensional spaces where vector magnitude is less significant. Combining triplet loss and cosine distance metrics can improve feature space representation resulting in the model's improved classification as well as enhancing robustness to illumination variations and other factors affecting flame appearance, leading to enhanced generalization to unseen joint distributions in modalities, and increasing intra-class variance. Note that in Eq.~\ref{cosine}, we define a novel form of cosine distance (denoted by $\mathcal{L}_{cosL}$)  inspired by the concept of triplet loss ( $\mathcal{L}_{TL}$).  It is computed based on the average similarity between the sample embedding's features ($e^{(i)}_j$) and the same feature of the sample's corresponding class prototype ($c^{p}_{y^{(i)}j}$) utilized by  Eq.~\ref{triplet}. In Eq.~\ref{triplet} and \ref{cosine}, the superscripts $p$ and $n$ denote the current samples' corresponding class prototype and the opposite class prototype respectively. 
At last, to force the representative to be close to the distribution of the clusters' embeddings, we apply another DML, called center loss (denoted by $\mathcal{L}_{CL}$ (attentive)). However, our experiments reveal that this unmodified version of $\mathcal{L}_{centL}$ is too dominant and hinders the positive contribution of $\mathcal{L}_{TL}$ and $\mathcal{L}_{cosL}$ in the learning process. As a result, the model will focus on discrimination in feature magnitudes, pushing representatives out of the cluster space to increase their difference magnitude. To address this concern, we integrate the deep attentive center loss ($\mathcal{L}_{CL}$), as proposed by \cite{farzaneh2021facial}. It is noteworthy that the adapted center loss in Eq.~\ref{mod_center} is employed to alleviate the prevailing constraints of the vanilla center loss ($a_{ij}$  represents feature attention weights). The attention mechanism enables each of the loss functions to scale features of the cluster representatives in distance calculation and thus, modifies the trajectory of the representative in the latent space, maximizing inter-class separation and minimizing intra-class separation. The calculation of $a^{TL}_{ij}, a^{CL}_{ij}, a^{CosL}_{ij} $ in Eq. ~\ref{triplet}, ~\ref{mod_center}, and ~\ref{cosine} is shown in the attention module in  (Fig \ref{main idea}) . In these equations, the non-attentive (vanilla) implementation is equivalent to setting all attention coefficients to 1 ($a^{TL}_{ij} = a^{CL}_{ij} =  a^{CosL}_{ij} = 1$). Moreover, all norms used are Euclidian norms. It should be noted that $a^{TL}_{ij}$, $a^{cosL}_{ij}$, $a^{CL}_{ij}$, $e^{(i)}_{j}$, and $ c^{p}_{y^{(i)}{j}}, c^{n}_{y^{(i)}{j}}$ (used in Eq. \ref{triplet},  \ref{mod_center}, \ref{cosine}, and later in Eq. \ref{der_to_Cp_TL}, and \ref{der_to_Cp_CL}), are all scalars, whereas $a^{cosL}_{i}$,  $e^{(i)}$, $c^{n}_{y^{(i)}}$, and $c^{p}_{y^{(i)}}$  (used in Eq. \ref{der_cos}) are vectors with length $c_{d}$, and the $"\odot"$ and  $"\cdot"$ symbols used in Eq.\ref{der_cos} denote element-wise multiplication and inner product operations respectively.

\setlength{\jot}{-0.1pt}
\begin{equation}
\label{DML_loss}
\mathcal{L}_{DML} = \,\gamma_1 \, \mathcal{L}_{TL} + \gamma_2 \, \mathcal{L}_{cosL} + \gamma_3 \, \mathcal{L}_{CL}, 
\end{equation}

\vspace{-10pt}
\begin{align}
\label{triplet}
\mathcal{L}_{TL} = \, \max \Big(0 \, ,& \, \frac{1}{2m} \sum_{i=0}^{m} \sum_{j=1}^{c_d} {[\, a^{TL}_{ij} \,(\,\| {(e^{(i)}_{j})} - (c^{p}_{y^{(i)}{j}}) \| _{2}^{2}} \nonumber\\ 
& -{\| ({e^{(i)}_{j}}) - (c^{n}_{y^{(i)}{j}}) \| _{2}^{2}} + \alpha)\,]\,\Big)
\end{align}

\vspace{-15pt}
\begin{align}
\label{mod_center}
\mathcal{L}_{CL} = \, & \, \frac{1}{2m} \sum_{i=0}^{m} \sum_{j=1}^{c_d} a^{CL}_{ij}  {\| {(e^{(i)}_{j})} - (c^{p}_{y^{(i)}{j}}) \| _{2}^{2}}\\ \nonumber
& st.~ ~ 0 < a_{ij} \leq 1 ~ ~ \forall j \mid j\in \{1,..., c_d\}
\end{align} 

\vspace{-15pt}
\begin{align}
\label{cosine}
\mathcal{L}_{cosL} = \, \, & - \, \frac{1}{2m} \sum_{i=0}^{m} \sum_{j=1}^{c_d}  a^{cosL}_{ij} \,\left[ \Big( \frac{(e^{(i)}_{j} c^{p}_{y^{(i)}{j}})}{\left\lVert (e^{(i)}) \right\rVert \big\lVert (c^{p}_{y^{(i)}}) \big\rVert} \right. \nonumber\\
-&\left. \frac{(e^{(i)}_{j}) (c^{n}_{y^{(i)}{j}})}{\left\lVert (e^{(i)}) \right\rVert \big\lVert (c^{n}_{y^{(i)}}) \big\rVert} \Big) \right] + 2,
\end{align}

\subsubsection{DML Metrics Derivatives}
Our proposed method (Alg.~\ref{proposed instruction}) trains the integral parameters of the loss functions, including the cluster representatives, in an end-to-end manner during supervised training using SGD. In this regard, Eq.~\ref{der_to_e_TL} -  Eq.~\ref{der_to_a_CosL} demonstrate the required closed-from derivatives for performing gradient descent. The closed-form derivatives of Eq. \ref{triplet} described in Eq. \ref{der_to_e_TL}, \ref{der_to_a_TL}, is required as the cluster representatives are getting updated every step and build the foundation for the inference phase. An important fact is that the computational graph cannot yield the derivatives of the loss functions concerning the the representative; thus closed-form derivatives are required for the moving-average centroid updates described in  \ref{der_to_Cp_CL}, \ref{der_to_Cp_TL}, and \ref{der_cos}. Parameter $k$ in Eq.~\ref{der_to_e_TL} is the number of samples that do not satisfy the margin defined in Eq.~\ref{triplet}, and so they contribute to learning, in other words, $k = \text{card}(\mathcal{K})$ where, $\mathcal{K}=\{x^{(i)}\,\, | \,\, \mathcal{L}_{TL}(x^{(i)}) > 0\}$. Inspired by \cite{farzaneh2021facial}, we follow a moving average strategy to update the class/cluster representatives, $C$. In this context, we divide the set $\mathcal{K}$ into two subsets of flame samples $\mathcal{K}_1$ and non-flame samples $\mathcal{K}_2$ given that, $\mathcal{K}_1 \cup \mathcal{K}_2 = \mathcal{K}$. In the same way, $\mathcal{K^{'}}=\{x^{(i)}\,\, | \,\, \mathcal{L}_{cos(\theta)}(x^{(i)}) > 0\}$, and $\mathcal{K}_1^{'}$, and $\mathcal{K}_2^{'}$  are defined as subsets of $\mathcal{K}^{'}$ that include flame and no-flame samples, respectively. Following the same notation, $\mathcal C_{TL}^{{K}_1}$ denotes the flame cluster representative, while $\mathcal C_{TL}^{{K}_2}$ denotes the no-flame cluster representative. It should be noted that both flame and no-flame samples affect the cluster representative gradient.



\noindent
\begin{minipage}{\linewidth}
    \noindent\rule{\linewidth}{0.7pt} 
    \textbf{Algorithm DML-Train (Alg.~\ref{proposed instruction}):} DML-aided Training \\
    \noindent\rule{\linewidth}{0.7pt} 

    \textbf{Inputs:}
    \begin{itemize}[itemsep=1pt, topsep=-3pt, label=\textbullet, font=\small, leftmargin=4ex]
        \item Batch of labeled cropped patches as training dataset, $D_t = \{ (x^{(i)}, y^{(i)}) \mid i = 1, 2, \ldots, N_t \}$,
        \item Initialized encoder-decoder network parameters ($\theta_{en}$, $\theta_{de}$, and $\theta_{cls}$),
        \item Reconstruction loss, DML, and softmax loss parameters ($\theta_{re}$, $\theta_{DML}$, $\theta_{s}$),
        \item Class (cluster) prototype, $c = \{ c_k \mid k = 1 , 2 ; \, c_{(k)} \in \mathbb{R}^{1 \times C_{d}} \}$,
        \item Hyperparameters ($\lambda_1$, $\lambda_2$, $\gamma_1$, $\alpha$),
        \item $t \,\,\,\,\, \longleftarrow \,\,\,\,\, 0$.
    \end{itemize}
    
    \vspace{0.2cm}
    \textbf{Output:}
    \vspace{0.1cm}

    \begin{itemize}[itemsep=1pt, topsep=-3pt, label=\textbullet, font=\small, leftmargin=4ex]
        \item Trained model and updated $\theta_{en}$, $\theta_{de}$, $\theta_{re}$, $\theta_{s}$, $\theta_{cls}$, and $C$.
    \end{itemize}
    \vspace{0.3cm}
    
    \begin{enumerate}[label=\arabic*., itemsep=1pt, topsep=-3pt, font=\small, leftmargin=*]
        \item[]\textbf{While} not converged \textbf{do}:
    
        \begin{enumerate}[label=\arabic*, leftmargin=5ex]
            \item Feed the input dataset minibatch $B_m = \{(x^{(i)},y^{(i)}) \mid i=0,1,...,m \}$ into the network.
            \item In feature space, compute the corresponding embeddings $e \in \mathbb{R}^{C_{f} \times 1 \times 1}$.
            \item Compute the CE loss utilizing Eq.~\ref{CE_loss}.
            \item Compute the AE loss utilizing Eq.~\ref{rec_loss}.
            \item Compute the DML loss utilizing Eq.~\ref{DML_loss} to Eq.~\ref{cosine}.
            \item Compute loss function gradient using Eq.~\ref{der_to_e_TL} to Eq.~\ref{der_to_a_CL} and chain rule.
            \item Compute $\Delta_C$ using Eq.~\ref{der_to_Cp_TL} to Eq.~\ref{der_to_Cp_CL}.
            \item $t \,\,\,\,\, \longleftarrow \,\,\,\,\, t+1$.
            \item $C^{t+1} \,\,\,\,\, \longleftarrow \,\,\,\,\, C^t-\alpha \Delta_C$.
        \end{enumerate}
        \item[]\textbf{End while}
    \end{enumerate}
    
   \vspace{1ex} 
    
    \hrule 
    \label{proposed instruction}
\vspace{3pt}
\end{minipage}

\begin{flalign}
\label{der_to_Cp_TL}
\Delta_{C_{TL}^{\mathcal{K}_l}}& = \frac{-1}{k}
\bigg( \sum_{x^{(i)} \in \mathcal{K}_l}^{} \,\, \sum_{j=1}^{c_d} {a^{TL}_{ij}  \big( {(e^{(i)}_{j})} - (c^{p}_{y^{(i)}{j}})} \big) - & \nonumber\\
& \sum_{x^{(i)} \in \mathcal{K}_q}^{} \,\sum_{j=1}^{c_d} {a^{TL}_{ij} \big( {(e^{(i)}_{j})} - (c^{n}_{y^{(i)}{j}})} \big)\bigg); &
\end{flalign}

\vspace{-10pt}
\begin{flushleft}
\begin{align}
\label{der_cos}
 \Delta_{C_{cosL}^{\mathcal{K}^{'}_1}} = \;\;\;\;\;& \nonumber\\
\frac{-1}{2k^{'}}  \bigg( \sum_{x^{(i)} \in \mathcal{K}_l^{'}}^{} & a^{cosL}_{i} \odot \Big(\frac{e^{(i)}}{\big\lVert e^{(i)}\big\rVert \big\lVert c^{p}_{y^{(i)}}\big\rVert} - \frac{(e^{(i)}\cdot c^{p}_{y^{(i)}})c^{p}_{y^{(i)}}}{\big\lVert e^{(i)}\big\rVert \big\lVert c^{p}_{y^{(i)}}\big\rVert ^3}\Big) \nonumber\\
- \sum_{x^{(i)} \in \mathcal{K}_q^{'}}^{} & a^{cosL}_{i} \odot \Big(\frac{e^{(i)}}{\big\lVert e^{(i)}\big\rVert \big\lVert c^{n}_{y^{(i)}}\big\rVert} - \frac{(e^{(i)}\cdot c^{n}_{y^{(i)}})c^{n}_{y^{(i)}}}{\big\lVert e^{(i)}\big\rVert \big\lVert c^{n}_{y^{(i)}}\big\rVert ^3}\Big) \bigg)
\end{align}
\end{flushleft}

\vspace{-15pt}
\begin{flushleft}
\begin{align}
\label{der_to_Cp_CL}
\Delta_{C_{CL}} = \frac{-1}{m}
\sum_{i=1}^{m} \,\, \sum_{j=1}^{c_d} a^{CL}_{ij} \big( {(e^{(i)}_{j})} - (c_{y^{(i)}{j}}) \big)
\end{align}
\end{flushleft}

where in Eq.~\ref{der_to_Cp_TL} and Eq.~\ref{der_cos}~; $(l, q) \in \left \{ (1, 2), (2, 1) \right \}$

\subsection{DML-guided Inference}

\begin{figure}[t]
\centerline{\includegraphics[width=0.8\columnwidth]{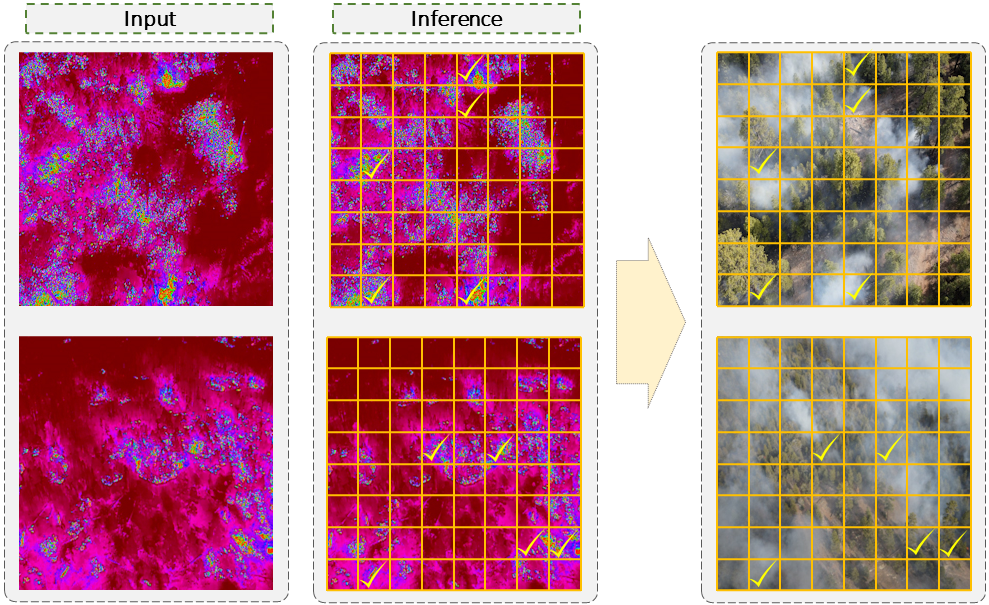}}
\captionsetup{justification=justified}
\caption{Obscured fire detection during inference. Only thermal patches are used for inference. The predicted labels correspond to the equivalent RGB patches and reveal potential obscured flames. }
\label{main result}
\vspace{-10pt}
\end{figure}

Once the Autoencoder is trained, the decoder is removed, leaving the classifier. The thermal domain cluster prototypes (Fig. \ref{main idea}) represent classes for 'flame' and 'no-flame' in the latent feature space. Back in training, the classifier had received the latent feature vector of a sample, while DML losses had made the embedding function generate more separable clusters and better prototypes. The unsupervised metrics used in supervised training exist to improve generalization for obscured flames with thermal patterns (where no ground truth is available). After training the optimal cluster representative (Alg.~\ref{proposed instruction}), the model predicts the class for obscured samples. The cluster representatives trained by the proposed DML approach set the prediction criteria (with fixed attention weights) and test samples are compared to each representative to make predictions. It should be noted that only thermal patches are used in the inference phase and the predicted labels for each thermal patch will eventually correspond to the equivalent RGB patch, revealing possible flames hindered by the smoke. (Fig. \ref{main result}).

\begin{figure}
    \centering
    \includegraphics[width=\linewidth]{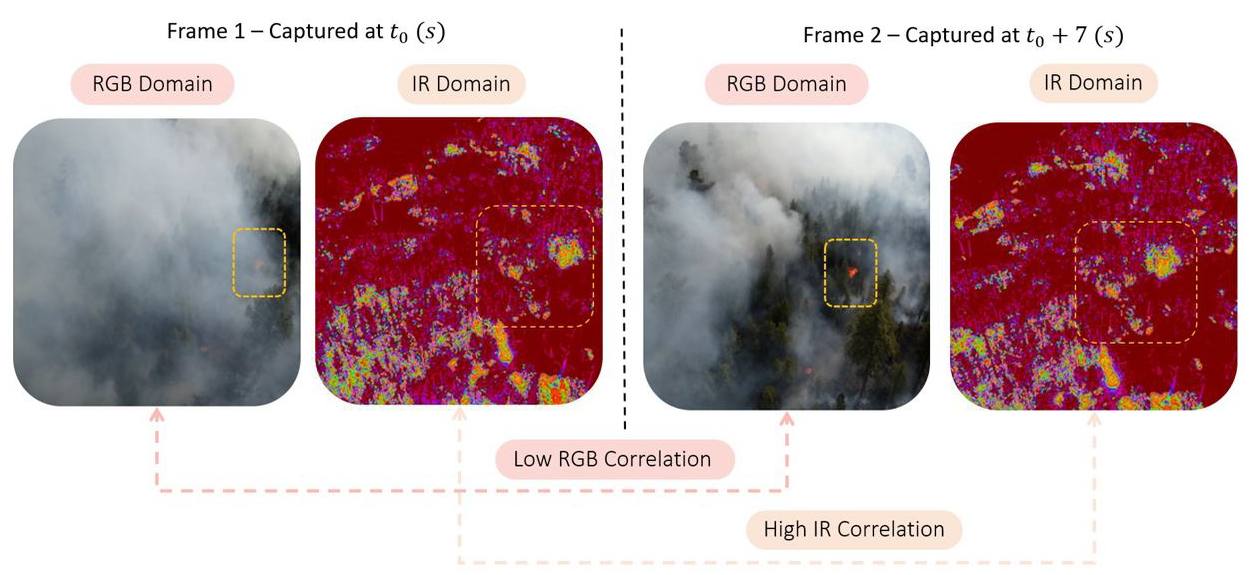}
    \caption{Thermal domain spatial distribution remains nearly unchanged after smoke obscures a flame patch and its surroundings.}
\vspace{-10pt}
\label{IR_Dist}
\end{figure}

As there are no patch labels to evaluate the model's performance on the test set, we use intra-class variability in the flame "class" as an unsupervised indicator to show how DML and attention make clusters as separable as possible by moving the prototypes and shaping the embedding function.  (Table. \ref{tab:Performance Metrics}) Higher separation suggests that RGB domain labels are good indicators of thermal domain features, indicating successful knowledge transfer. It should be noted that successful knowledge transfer relies on the fact that obscured and unobscured thermal domain representations of flame samples are significantly correlated. Fig. \ref{IR_Dist} shows two frames captured 7 seconds apart, from FLAME2, showing the same flame, but with different smoke occlusion density and patterns around the flame.  RGB representation looks very different due to smoke occlusion, yet IR domain greatly preserves the distribution, showing class-specific feature transparency through the smoke. Additionally, RGB representations of the obscured images look pretty similar (low inter-class variance) and class-specific features are obscured by smoke. Thus, using RGB along IR for inference acts as noise, making class separation more difficult. This is the motivation behind using only thermal domain patches for inference. \subsection{Local Features Extraction (LFE) Mechanism}  To capture distinctive features from specific subareas of an image, both the thermal image, denoted as $x^{(i)} $, and the masked thermal image, represented as $x^{(i)}_{m} = x^{(i)} \odot m^{(i)}$, (where $\odot$ signifies the Hadamard product (element-wise multiplication) and $m^{(i)}$ is the corresponding mask generated by Alg.~\ref{unobscured flame segmentation}), are fed into two U-Nets with shared parameters. Subsequently, the local and global features are aggregated within the latent space as illustrated in Fig. \ref{main idea}. This approach enables the model to focus on masked subareas containing intricate patterns and textures associated with flame spots. Given that wildfires' early-stage flames typically manifest as dispersed spots, this methodology facilitates a detailed understanding of such regions through targeted analysis of the masked portions.\vspace{0.2cm}  

\vspace{-0.2cm}
\section{Performance Evaluation} 

 In this section, first, we will cover the data selection procedure, dataset partitioning, pre-processing, and training. Next, we will discuss the results obtained from training, validation, hyper-parameter tuning, and testing in terms of supervised and unsupervised metrics. \vspace{-0.4cm}

\subsection{Dataset}
The FLAME2, \cite{fule2022flame},  and FLAME3 \footnote[1]{FLAME3 dataset is collected by a team of universities, US Forest Services, and CALFIRE, and is not publicly available.}  datasets have been used for evaluation. FLAME2 includes seven side-by-side infrared and visible spectrum video pairs captured by drones from open-canopy fires in Northern Arizona in 2021. Derived from the video pairs, a set labeled original resolution RGB/Thermal frame pairs and a set of 53'000 labeled 254p x 254p RGB/Thermal frame pairs are provided. FLAME 2 provides two labels of fire/no-fire and smoke/no-smoke for every image. Experts have assigned fire labels based on visual fire indicators in the RGB domain. Moreover, they have assigned 'smoke' and 'no-smoke' labels to every RGB image, resulting in 14'760 smoke-free frames. The RGB frames are the source of visual labeling, while thermal images are provided along them to extract the thermal features of the supervised set. The corresponding details for FLAME2 and FLAME3 and the experimental settings are shown below, in Table \ref{tab:experiment}.

\label{appendix:experiment settings}

\begin{table}[h]
  \centering
  \caption{Dataset Details and Experiment Settings}
  \label{tab:experiment}
  
  \renewcommand{\arraystretch}{1.3} 

  \begin{tabular}{|c|c|c|} \hline  
    
    \textbf{Dataset Details} & \textbf{\textit{FLAME 2}} & \textbf{\textit{FLAME 3\text{$^*$}}} \\ \hline  
    
    \# RGB/Thermal Img. Pairs & $53k$ & $20k$ \\ \hline  
    
    \textbf{Experiment Settings} & \textbf{\textit{FLAME 2}} & \textbf{\textit{FLAME 3}} \\ \hline  
    
    \# Total Samples (Train) & $4700^**$ & $880^{***}$ \\ \hline  
    \# Total Samples (Test) & 1170 & 220  \\ \hline  
    \# Batches/Sample & $64$ &  $64$\\ \hline  
    \# Original Sample Resolution  & $256\times256$ &  $512\times640$\\ \hline  
    \# Input Resolution  & $256\times256$ &  $256\times256$\\ \hline  
    \# Batch Resolution & $32\times32$ &   $32\times32$ \\ \hline

  \end{tabular}
\end{table}

*-  Flame 3 stats are not final.

**- A portion of Flame2 samples are utilized during training and testing.

***- The Shoetank's fire samples of Flame3 are used.


\begin{figure}[htb]
\centerline{\includegraphics[width=0.93\columnwidth]{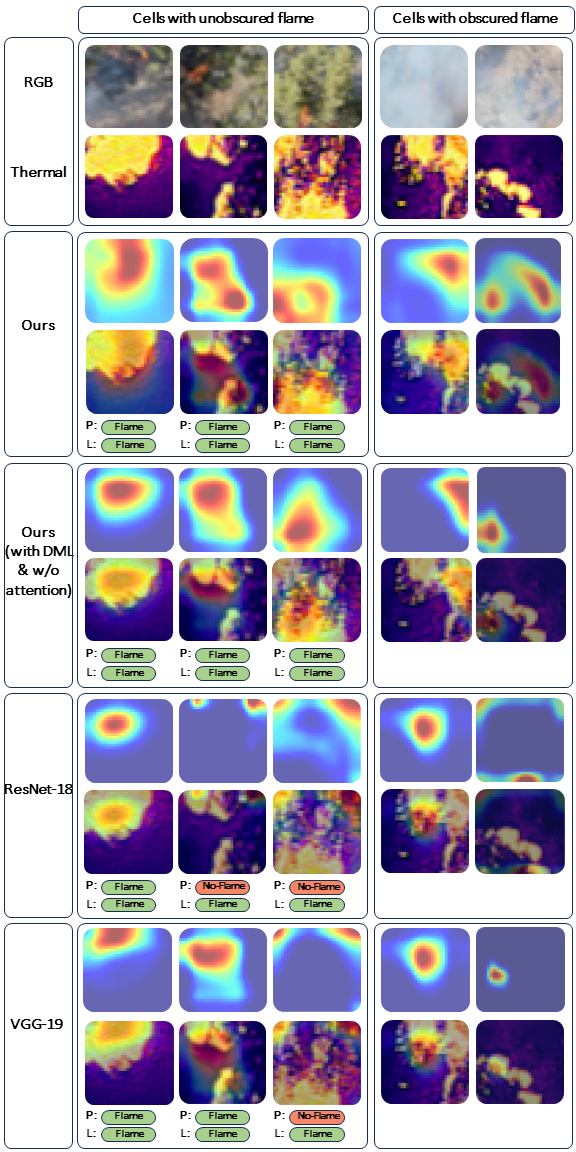}}
\captionsetup{justification=justified}
\caption{Visual performance comparison of the proposed method with baseline models. The gradient magnitude of the last convolutional layer, based on the idea introduced in \cite{selvaraju2017grad} (Grad-CAM) is upsampled to the image size and stacked on the sample image. P and L denote prediction and label, respectively. 
}
\label{visual results}
\vspace{-15pt}
\end{figure}

\subsection{Implementation}
\label{implementation}

In our work, we used $4700$ frames, or one-third of the smoke-free samples of the labeled RGB/Thermal image pairs in FLAME2, along with 1200 samples of FLAME3 to diversify the dataset and avoid redundant frames. We start preprocessing by upsampling frames to $256p \times 256p$ resolution. Next, we divide each image into 64 $32p \times 32p$ patches. This applies to RGB and thermal images. After that, we segment fire-containing RGB images using Alg.~\ref{unobscured flame segmentation}. The segmentation parameters were manually set to $\left \{\alpha = 0.1, \beta = 0.47, \delta = 0.14 \right \}$. The annotation accuracy on smoke-free patches is compared to pixel-annotated patches labeled by an expert. The segmentation is compared to a human-expert perfect baseline performed on $8640$ patches originating from $135$ images, through which the segmentation algorithm was able to identify $ 81\%$ of the patches annotated as 'flame' by the expert. The gap is assumed to be mainly due to the presence of obscure and partially ignited patches which are seen by the expert and rejected by the RGB mask. While not achieving human expert-level performance, Alg. \ref{unobscured flame segmentation} offers an automated segmentation for large-scale datasets. The comparative analysis between manual and automated labeling is present in \ref{annotation analysis} featuring Table \ref{tab:confusionmatrix}. Next, annotated thermal patches are over-sampled to create a balanced dataset of 'flame' and 'no-flame' patches. Data augmentation is performed by random flipping vertically and horizontally and normalizing. Our model's encoder, decoder, and classifier were trained using the procedure shown in Fig. \ref{main idea}. The FLAME2 dataset was used for 20 epochs of training with an initial learning rate of $0.01$ for both $\mathcal{L}_{BCE}$ and $\mathcal{L}_{rec}$. The learning rate decreased by $0.1$ at epochs $10$ and $18$ for $\mathcal{L}_{BCE}$ and $\mathcal{L}_{rec}$, respectively. Each batch of thermal patches had $512$ images to train the model. Empirical tuning yielded the following training hyperparameter values: $\lambda = 1$, $\gamma_1 = 0.01$, $\gamma_2 = 0.0001$, and $\gamma_3 = 0.01$. In addition, we set DML learning rates as follows: $\alpha_{TCL} = 0.5$, $\alpha_{CL} = 0.5$, and $\alpha_{CosL} = 0.1$. Our models are trained using the PyTorch deep learning framework on an NVIDIA 3080 GPU. 

\begin{table}[h]
\centering
\captionsetup{justification=justified}
\caption{Performance comparison against other studies. (ICV: Intra-Class Variance, FNR: False Negative Rate). Metrics for Flame 2 and 3 are reported on the left and right respectively. }\vspace{0pt}
\label{tab:Performance Metrics}

\renewcommand{\arraystretch}{.05}
\begin{tabular}
{>{\centering\arraybackslash}p{0.7cm}
 >{\centering\arraybackslash}p{2.3cm}
 >{\centering\arraybackslash}p{0.9cm}
 >{\centering\arraybackslash}p{0.8cm}
 >{\centering\arraybackslash}p{1.1cm} 
  }
\toprule
{\fontsize{6}{6}\selectfont { \rule{-5pt}{0ex}
\begin{center}
\textbf{Dataset}
\end{center}
}}
&
{\fontsize{6}{6}\selectfont { \rule{-5pt}{0ex}
\begin{center}
\textbf{Model}
\end{center}
}}
&
{\fontsize{6}{6}\selectfont { \rule{-5pt}{0ex}
\begin{center}
\textbf{Accuracy}
\end{center}
}}
&
{\fontsize{6}{6}\selectfont { \rule{-8pt}{0ex}
\begin{center}
\textbf{False Negative}
\end{center}
}}
&

{\fontsize{6}{6}\selectfont { \rule{-8pt}{0ex}
\begin{center}
\textbf{Intra-Class Variance}
\end{center}
}}

\\[-1.5ex]
 \midrule


&
{\fontsize{7}{6}\selectfont {\rule{-3pt}{0ex}
ResNet18
}}
&
{\fontsize{7}{6}\selectfont {\rule{-5pt}{0ex}
93.70
}}
&
{\fontsize{7}{6}\selectfont {\rule{-5pt}{0ex}
17.00
}}
 &

{\fontsize{7}{6}\selectfont {\rule{-5pt}{0ex}
48.70
}}
\\[+1.5ex]

&
{\fontsize{7}{6}\selectfont {\rule{-3pt}{0ex}
VGG19
}}
&
{\fontsize{7}{6}\selectfont {\rule{-5pt}{0ex}
92.90
}}
&
{\fontsize{7}{6}\selectfont {\rule{-5pt}{0ex}
17.40
}}
 &

{\fontsize{7}{6}\selectfont {\rule{-5pt}{0ex}
49.70
}}
 
\\[-1ex]

&

{\fontsize{7}{6}\selectfont {\rule{-3pt}{0ex}
ShRe Xception \cite{zachariadis2023}
}}
&
{\fontsize{7}{6}\selectfont {\rule{-5pt}{0ex}
93.50
}}
&
{\fontsize{7}{6}\selectfont {\rule{-5pt}{0ex}
18.60
}}
 &

{\fontsize{7}{6}\selectfont {\rule{-5pt}{0ex}
50.30
}}
 
\\[-1ex]

\multirow{1}{*}{%
    \fontsize{7}{6}\selectfont
    \centering
    \textbf{Flame2}
}
&
{\fontsize{7}{6}\selectfont {\rule{-3pt}{0ex}
Ensemble model \cite{ghali2022deep}
}}
&
{\fontsize{7}{6}\selectfont {\rule{-5pt}{0ex}
94.20
}}
&
{\fontsize{7}{6}\selectfont {\rule{-5pt}{0ex}
14.10
}}
 &

{\fontsize{7}{6}\selectfont {\rule{-5pt}{0ex}
46.60
}}
 
\\[-1ex]
&

{\fontsize{7}{6}\selectfont {\rule{-3pt}{0ex}
CT-Fire \cite{ghali2023ct}
}}
&
{\fontsize{7}{6}\selectfont {\rule{-5pt}{0ex}
94.80
}}
&
{\fontsize{7}{6}\selectfont {\rule{-5pt}{0ex}
13.50
}}
 &

{\fontsize{7}{6}\selectfont {\rule{-5pt}{0ex}
43.00 
}}
 
\\[-1ex]
&

{\fontsize{7}{6}\selectfont { \rule{-3pt}{0ex}
Ours
}}
&
{\fontsize{7}{6}\selectfont {\rule{-5pt}{0ex}
\textbf{97.80}
}}
&
{\fontsize{7}{6}\selectfont {\rule{-5pt}{0ex}
\textbf{11.60}
}}
&
{\fontsize{7}{6}\selectfont {\rule{-5pt}{0ex}
\textbf{39.70}
}}



\\[-1ex]
 \midrule

&
{\fontsize{7}{6}\selectfont {\rule{-3pt}{0ex}
ResNet18
}}
&
{\fontsize{7}{6}\selectfont {\rule{-5pt}{0ex}
81.40
}}
&
{\fontsize{7}{6}\selectfont {\rule{-5pt}{0ex}
19.10
}}
 &

{\fontsize{7}{6}\selectfont {\rule{-5pt}{0ex}
50.90
}}
\\[+1.5ex]

&
{\fontsize{7}{6}\selectfont {\rule{-3pt}{0ex}
VGG19
}}
&
{\fontsize{7}{6}\selectfont {\rule{-5pt}{0ex}
80.50
}}
&
{\fontsize{7}{6}\selectfont {\rule{-5pt}{0ex}
18.80
}}
 &

{\fontsize{7}{6}\selectfont {\rule{-5pt}{0ex}
49.20
}}
 
\\[-1ex]

&

{\fontsize{7}{6}\selectfont {\rule{-3pt}{0ex}
ShRe Xception \cite{zachariadis2023}
}}
&
{\fontsize{7}{6}\selectfont {\rule{-5pt}{0ex}
81.90
}}
&
{\fontsize{7}{6}\selectfont {\rule{-5pt}{0ex}
18.20
}}
 &

{\fontsize{7}{6}\selectfont {\rule{-5pt}{0ex}
49.00
}}
 
\\[-1ex]

\multirow{1}{*}{%
    \fontsize{7}{6}\selectfont
    \centering
    \textbf{Flame3}
}
&
{\fontsize{7}{6}\selectfont {\rule{-3pt}{0ex}
Ensemble model \cite{ghali2022deep}
}}
&
{\fontsize{7}{6}\selectfont {\rule{-5pt}{0ex}
83.80
}}
&
{\fontsize{7}{6}\selectfont {\rule{-5pt}{0ex}
16.80
}}
 &

{\fontsize{7}{6}\selectfont {\rule{-5pt}{0ex}
47.40
}}
 
\\[-1ex]
&

{\fontsize{7}{6}\selectfont {\rule{-3pt}{0ex}
CT-Fire \cite{ghali2023ct}
}}
&
{\fontsize{7}{6}\selectfont {\rule{-5pt}{0ex}
85.10
}}
&
{\fontsize{7}{6}\selectfont {\rule{-5pt}{0ex}
16.90
}}
 &

{\fontsize{7}{6}\selectfont {\rule{-5pt}{0ex}
45.10
}}
 
\\[-1ex]
&

{\fontsize{7}{6}\selectfont { \rule{-3pt}{0ex}
Ours
}}
&
{\fontsize{7}{6}\selectfont {\rule{-5pt}{0ex}
\textbf{88.90}
}}
&
{\fontsize{7}{6}\selectfont {\rule{-5pt}{0ex}
\textbf{14.30}
}}
&
{\fontsize{7}{6}\selectfont {\rule{-5pt}{0ex}
\textbf{41.20}
}}

\\[-1ex]



\bottomrule
\end{tabular}
\vspace{-10pt}
\end{table}

\begin{table*}[h]
\centering
\captionsetup{justification=justified}
\caption{ Ablation Study, binary flame detection acc. for LFE and Attention, and Triplet Cosine and Cosine Losses} 1. Both adding DML and LFE show accuracy improvement, 2. Triplet and Cosine Losses improve DML after adding attentive CL
\label{tab:AchievementComparison}
\renewcommand{\arraystretch}{.5}
\begin{tabular}
{>{\centering\arraybackslash}p{1.5cm}
 >{\centering\arraybackslash}p{4cm}
 >{\centering\arraybackslash}p{1.5cm}
 >{\centering\arraybackslash}p{1.5cm}
 >{\centering\arraybackslash}p{1.5cm}
 >{\centering\arraybackslash}p{3cm} 
  }
\toprule
\multirow{3}{*}{\parbox{1cm}
{\fontsize{8}{7}\selectfont { \rule{0pt}{0ex}
\begin{center}
\textbf{Dataset}
\end{center}
}}}
&
 \multirow{3}{*}{\parbox{4cm}
{\fontsize{8}{7}\selectfont { \rule{0pt}{0ex}
\begin{center}
\textbf{Detection Method}
\end{center}
}}}
&
\multirow{3}{*}{\parbox{1.5cm}
{\fontsize{8}{7}\selectfont { \rule{0pt}{0ex}
\begin{center}
\textbf{No DML}
\end{center}
}}}
&
\multicolumn{3}{c}{\multirow{2}{*}{\parbox{2cm}
{\fontsize{8}{7}\selectfont { \rule{0pt}{0ex}
\textbf{DML Ablation}
}}}}
\\
& & & \multicolumn{3}{c}{} \\ \cmidrule{4-6} 
& & &
{\fontsize{8}{7}\selectfont { \rule{0pt}{0ex}
\textbf{CL}
}}
&
{\fontsize{8}{7}\selectfont { \rule{0pt}{0ex}
\textbf{CL \& TCL}
}}
&
{\fontsize{8}{7}\selectfont { \rule{0pt}{0ex}
\textbf{CL \& TCL \& CosL}
}}

\\
[-1.5ex]
 \midrule
 

\multirow{3}{*}{\parbox{1cm}
{\fontsize{7}{6}\selectfont 
{\rule{-3pt}{0ex}
\begin{center}
\textbf{Flame2}
\end{center}
}}}
&
{\fontsize{7}{6}\selectfont { \rule{0pt}{0ex}
Baseline
}}
&
{\fontsize{7}{6}\selectfont {\rule{-5pt}{0ex}
93.40
}}
&
{\fontsize{7}{6}\selectfont {\rule{-5pt}{0ex}
95.90
}}
&
{\fontsize{7}{6}\selectfont {\rule{-5pt}{0ex}
95.70
}}
 &
{\fontsize{7}{6}\selectfont {\rule{-5pt}{0ex}
95.50
}}

\\[-1ex]

&
{\fontsize{7}{6}\selectfont {\rule{-3pt}{0ex}
Baseline + LFE
}}
&
{\fontsize{7}{6}\selectfont {\rule{-5pt}{0ex}
95.00
}}
&
{\fontsize{7}{6}\selectfont {\rule{-5pt}{0ex}
96.10
}}
 &

{\fontsize{7}{6}\selectfont {\rule{-5pt}{0ex}
96.10
}}
 &
{\fontsize{7}{6}\selectfont {\rule{-5pt}{0ex}
95.90
}}

\\[-1ex]
&
{\fontsize{7}{6}\selectfont {\rule{-3pt}{0ex}
Baseline + LFE + Attention
}}
&
{\fontsize{7}{6}\selectfont {\rule{-5pt}{0ex}
Not Applicable
}}
&
{\fontsize{7}{6}\selectfont {\rule{-5pt}{0ex}
97.20
}}
 &

{\fontsize{7}{6}\selectfont {\rule{-5pt}{0ex}
97.50
}}
 &
{\fontsize{7}{6}\selectfont {\rule{-5pt}{0ex}
\textbf{97.80}
}}

\\[-1.5ex]

 \midrule
 

\multirow{3}{*}{\parbox{1cm}
{\fontsize{7}{6}\selectfont 
{\rule{-3pt}{0ex}
\begin{center}
\textbf{Flame3}
\end{center}
}}}
&
{\fontsize{7}{6}\selectfont { \rule{0pt}{0ex}
Baseline
}}
&
{\fontsize{7}{6}\selectfont {\rule{-5pt}{0ex}
81.9
}}
&
{\fontsize{7}{6}\selectfont {\rule{-5pt}{0ex}
84.40
}}
&
{\fontsize{7}{6}\selectfont {\rule{-5pt}{0ex}
85.10
}}
 &
{\fontsize{7}{6}\selectfont {\rule{-5pt}{0ex}
84.70
}}

\\[-1ex]

&
{\fontsize{7}{6}\selectfont {\rule{-3pt}{0ex}
baseline + LFE
}}
&
{\fontsize{7}{6}\selectfont {\rule{-5pt}{0ex}
83.80
}}
&
{\fontsize{7}{6}\selectfont {\rule{-5pt}{0ex}
86.00
}}
 &

{\fontsize{7}{6}\selectfont {\rule{-5pt}{0ex}
85.90
}}
 &
{\fontsize{7}{6}\selectfont {\rule{-5pt}{0ex}
85.80
}}

\\[-1ex]
&
{\fontsize{7}{6}\selectfont {\rule{-3pt}{0ex}
baseline + LFE + attention
}}
&
{\fontsize{7}{6}\selectfont {\rule{-5pt}{0ex}
Not Applicable
}}
&
{\fontsize{7}{6}\selectfont {\rule{-5pt}{0ex}
88.10
}}
 &

{\fontsize{7}{6}\selectfont {\rule{-5pt}{0ex}
88.80
}}
 &
{\fontsize{7}{6}\selectfont {\rule{-5pt}{0ex}
\textbf{88.90}
}}

\\[-1.5ex]
\bottomrule
\end{tabular}
\vspace{-10pt}
\end{table*}

\begin{figure}[htb]
    \centering
    \begin{subfigure}[t]{0.47\textwidth}
        \centering
        \includegraphics[width=\linewidth]{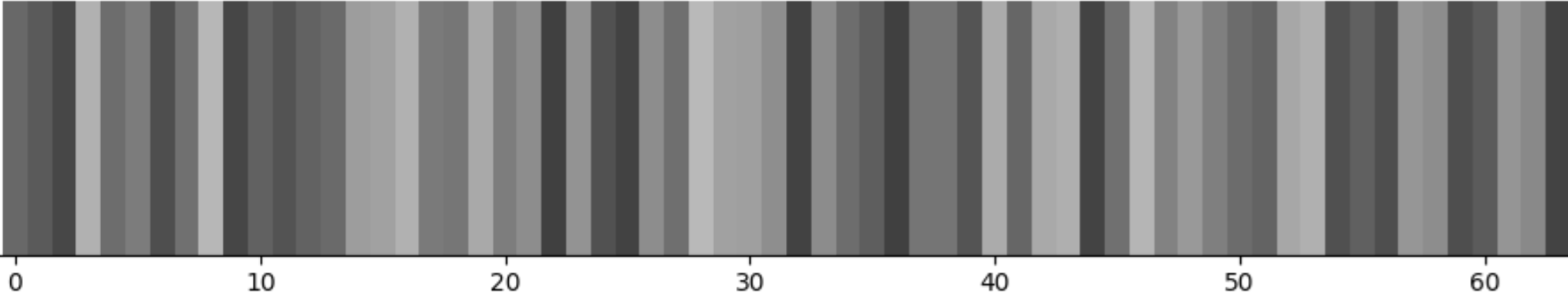}
        \caption{}
        \label{cos_att}
        \vspace{-0.25ex} 
    \end{subfigure}
    \hfill
    \begin{subfigure}[t]{0.47\textwidth}
        \centering
        \includegraphics[width=\linewidth]{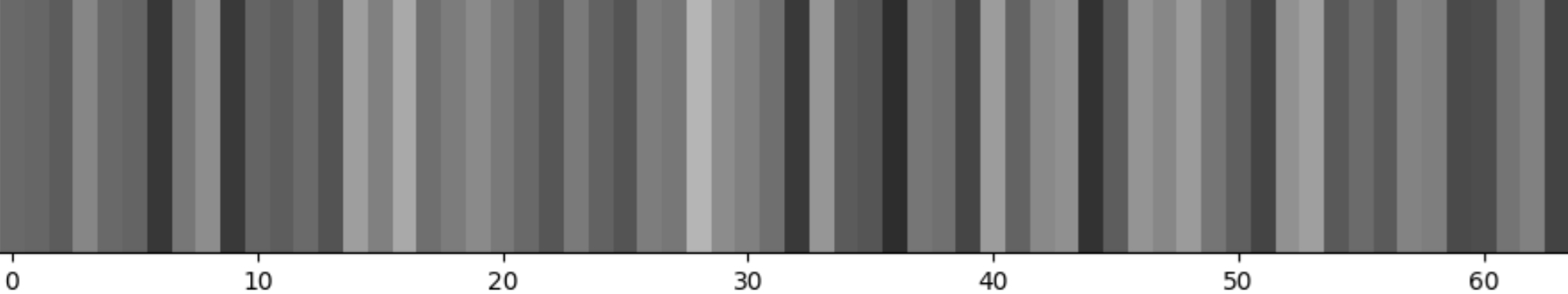}
        \caption{}
        \label{triplet_att}
        \vspace{-0.25ex} 
    \end{subfigure}
    \hfill
    \begin{subfigure}[t]{0.47\textwidth}
        \centering
        \includegraphics[width=\linewidth]{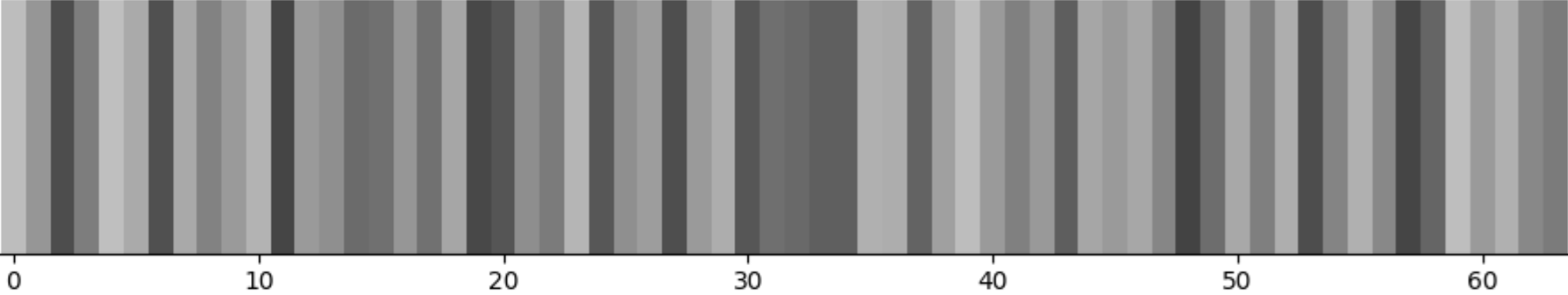}
        \caption{}
        \label{center_att}
    \end{subfigure}
    \captionsetup{justification=justified}
    \caption{Comparison of attention values as indicators for including and excluding features' contribution in DML losses value, Eq.~\ref{triplet} to Eq.~\ref{mod_center}, and in the backpropagation process, Eq.~\ref{der_to_e_TL} to Eq~\ref{der_to_a_CosL}: (a) Cosine-based loss attention, (b) Triplet loss attention, and (c) Center loss attention. The weight sparsity of the center loss (e.g. features 9, 36, 45, 64) shows the attention mechanism adaptively regularizes the contribution of every extracted latent feature in the loss functions, guiding the model towards the optimum in the parameter space (instead of scaling loss function derivatives uniformly with the hyper-tuning parameters and causing center-loss domination}.
    \label{conv}
\vspace{-10pt}
\end{figure}


\begin{figure}[h]
\centering
 \begin{tabular}{cc}
  \includegraphics[width=0.22\textwidth]{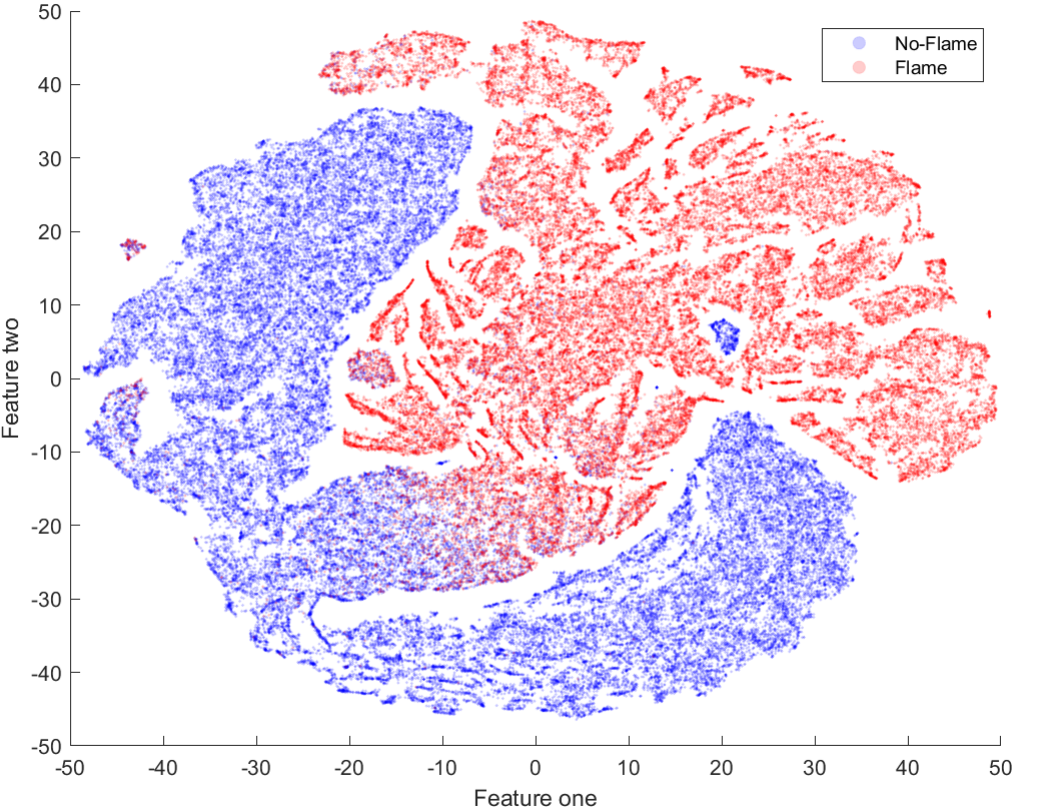}
  \label{fig:latent_scaterring_a}& \includegraphics[width=0.22\textwidth]{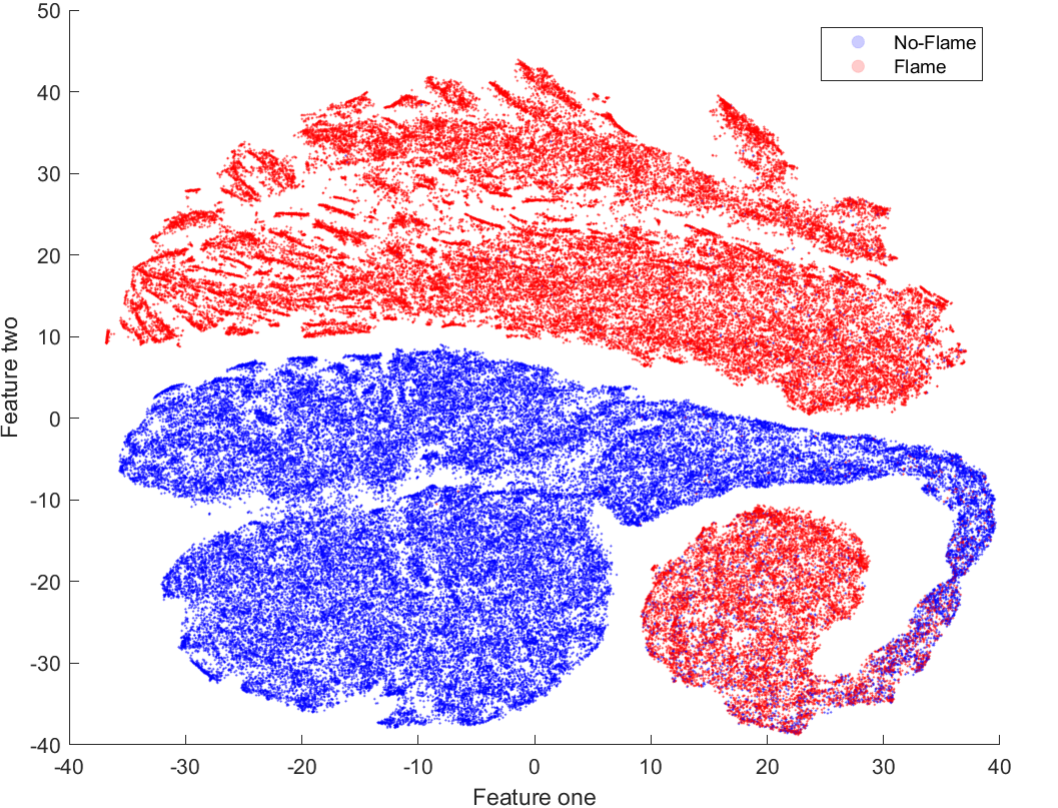} 
  \label{fig:latent_scaterring_b}\\
  (a) & (b) \\
\end{tabular}
\caption{2D demonstration of the clustered latent feature space. a. Baseline latent scattering and b. DML-guided latent scattering. Dimension reduction via t-SNE approach}
\end{figure}

\subsection{Latent Scattering Demonstration} \vspace{-0.1cm}

This study reduces latent space representation dimensionality using t-SNE (t-Distributed Stochastic Neighbor Embedding). The model is guided to the optimal subspace by the DML losses. The 2D plots in Fig. 5 
illustrate the effects of the proposed modules. The DML-guided model sub-figure shows significant improvement over Non-DML baseline model. 
Figure \ref{main idea} highlights misclassifications in some patches due to challenges posed by the non-optimal annotation mechanism in Alg.~\ref{unobscured flame segmentation}, making the discrimination between flame and no-flame classes in RGB space challenging. Misclassifications in flame patches arise from the lack of temperature calibration in the FLAME2 dataset, resulting in subclusters, the similarity between smoke and flame thermal distributions, and temperature gradient localized wind being generated and interacting with smoke patterns, moving them to areas far from the flame. 
\vspace{-5pt}


 \subsection{Numerical Results}
 In this section we present a thorough numerical analysis to: 1. Compare our model with previously proposed models for flame detection as well as VGG19 and ResNet18 backbones adapted for flame detection.  2. Evaluate how DML, LFE and Attention aid in boosting the classification accuracy as well as the impact of adding Triplet and Cosine Losses to  Center Loss before and after attention. \vspace{0.15cm} \\
 \textbf{Comparative Analysis:} As seen in Table \ref{tab:Performance Metrics}, our model achieves higher accuracy, and less false negative rate (miss) compared to all other backbones, including VGG-19, ResNet-18, and three other works specified for flame detection. Our model also achieves a lower intra-class variance showing the compactness of the latent clusters as an unsupervised metric. \vspace{0.2cm}\\
 \textbf{Ablation Study:} Table \ref{tab:AchievementComparison} shows an ablation study for DML, LFE, and Attention as well as Triplet and Cosine Losses. Samples acquiring less than 90\% confidence in one of the two classes are omitted from the prediction. For the baseline, the cluster representatives are calculated by averaging all feature vectors in the latent space. According to Table \ref{tab:AchievementComparison}, DML improves the binary classification accuracy. After adding the attention mechanism to overcome the center loss domination problem,  each latent feature contribution towards the proposed loss functions is balanced, demonstrating another 1.9\% and 3.1\%  accuracy improvement compared to the non-attentive version, for FLAME2 and FLAME3 respectively. Lower accuracy on the latter may be due to more diverse samples with variety in flame type, illumination, vegetation, etc. \vspace{-0.2cm}
 \subsection{Interpretabtiliy}
 The attentive classifier not only demonstrates favorable results but also how it has achieved them. With the aid of two main tools presented in this section, the contribution of all DML losses and global feature extraction is showcased.
\vspace{0.2cm}
\begin{itemize}
    \item \textbf{Grad-CAM:} As seen in Fig. \ref{visual results}, in the baseline models, according to the Grad-CAM \cite{selvaraju2017grad}, most of the learning is focused on high-intensity regions in the thermal domain, whereas most flame features in our model, especially with attention, are non-intensity based flame features, covering around the flame. This aligns with global feature discovery and acts as a solution to the lack of absolute thermal reference points decreasing false positives.
    \vspace{0.1cm}
    \item \textbf{Attention Map:} Fig. \ref{cos_att} to \ref{center_att} depict how each of the 64 features contributes to the aggregated loss gradient. The intensity of each band within the map corresponds to the attention weight of the feature associated with that band. The sparsity of features in the center loss attention map (lower weights - \ref{center_att}) shows the attention mechanism successfully overcomes the domination problem and refines the model towards capturing non-magnitude features in the latent space, provoking the cosine and triplet loss, which appear quite correlated (\ref{cos_att}, \ref{triplet_att}), in the learning process.  
\end{itemize}
 \vspace{-0.2cm}


\begin{figure}[t]
\centerline{\includegraphics[width=0.95\columnwidth]{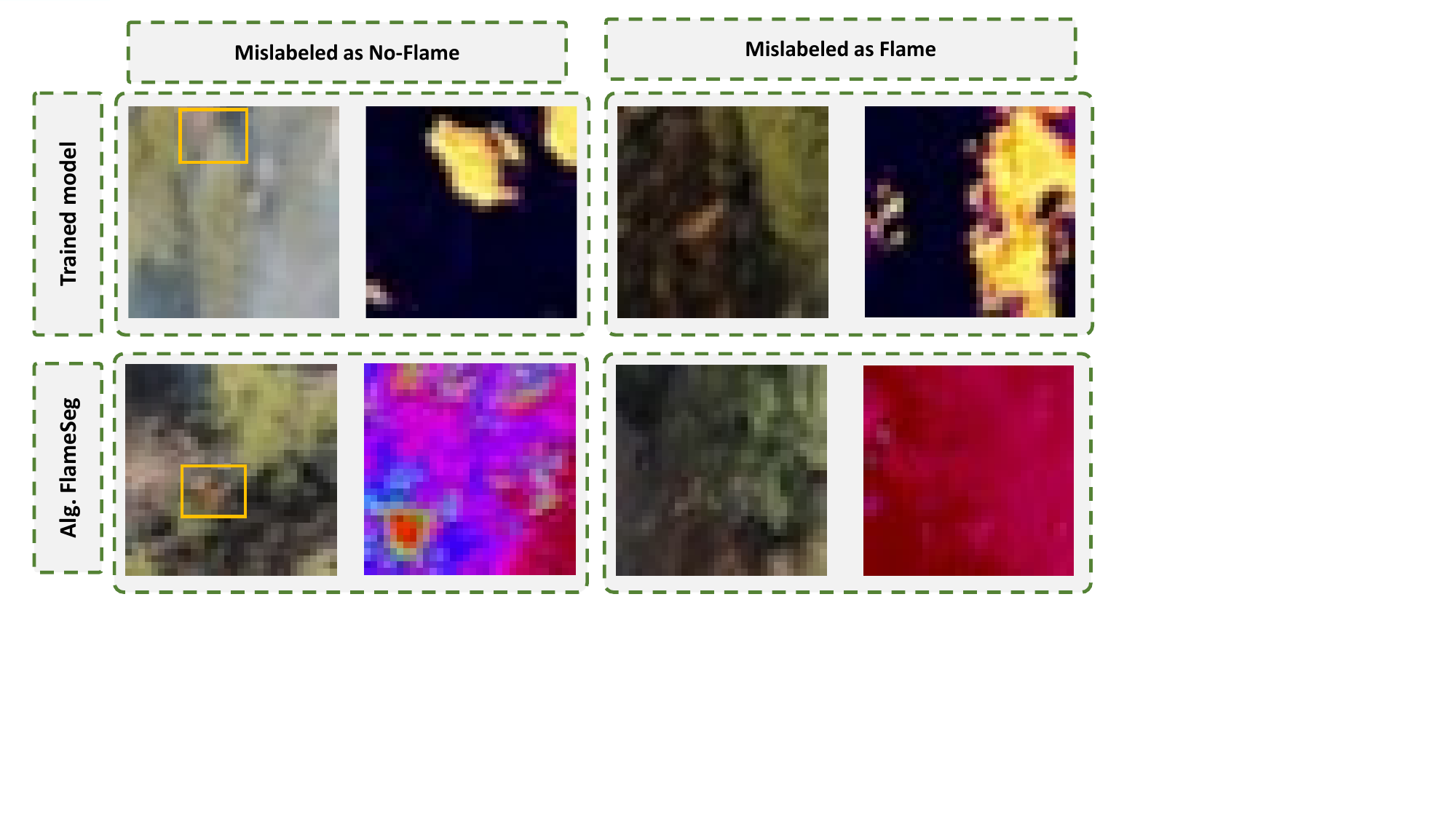}}
\captionsetup{justification=justified}
\caption{Classified and misclassified patches during validation. }
\label{mislabled_samples}
\end{figure}

\section{Conclusion and Future Work}

FlameFinder offers a comprehensive solution encompassing segmentation, DML-guided classification, and aided attention, leveraging the power of DML as a significant driving force for detecting obscured flames. The results achieved from the predictions of the DML-based model demonstrate favorable performance in terms of separating flames in both unobscured and obscured cases. Here, the attention mechanism used in DML, balances the utilization of features in separating latent clusters. As the proposed DML-based model receives segmentation masks as labels, it can be generalized to other application domains. The proposed model can also be utilized for other domain-adaptation tasks on multi-modal data by modifying the segmentation masks or proposing other application-specific annotation mechanisms. Despite DML's significant role in constructing better cluster representatives and compressing latent features around them, some samples may still be misclassified. As seen in Fig. \ref{mislabled_samples}, false positives are mostly cases with high thermal activity originating from smoke, neighboring flames, and other thermal sources that may be misclassified as flames. False negatives are mostly small flames (compared to the patch size), corresponding to early thermal patterns with different shapes than larger flames.  In general, the leading causes of mis-classification is assumed to include the lack of thermal reference points, similar thermal distributions in smoke and flame, thermal noise originating from temperature gradients of superimposed/interfered smoke layers in adjacent ignited areas, and lastly, the premature thermal distribution of early flames resulting in outliers relative to larger flames. These limitations open up new research directions in many fields, such as multi-modal detection, thermal imagery, and representation learning, as well as escalating the necessity of a more challenging dataset suited for generalization to real-world scenarios.



\newpage
\bibliographystyle{ieeetr}
\bibliography{main}


 





\vfill

\end{document}